\documentclass{article}

\usepackage[final]{neurips_data_2024}

\usepackage[utf8]{inputenc} 
\usepackage[T1]{fontenc}    
\usepackage{hyperref}       
\usepackage{url}            
\usepackage{booktabs}       
\usepackage{amsfonts}       
\usepackage{nicefrac}       
\usepackage{microtype}      
\usepackage{listings}
\usepackage{xcolor}         
\usepackage[pdftex]{graphicx}
\usepackage{subfig}
\usepackage{balance}
\usepackage{comment}
\usepackage{authblk}
\usepackage{xcolor}
\usepackage{xspace}
\usepackage{wrapfig}
\usepackage{caption}
\usepackage{tabularx}
\usepackage{wrapfig}

\newcommand{\emgtwoqwerty}{\textit{emg2qwerty}\xspace}

\setcitestyle{round}

\title{emg2qwerty: A Large Dataset with Baselines for Touch Typing using Surface Electromyography}

\author{Viswanath Sivakumar\thanks{Correspondence: viswanath@meta.com} }
\author{Jeffrey Seely\thanks{Work done while at Meta.} }
\author{Alan Du}
\author{Sean R Bittner}
\author{Adam Berenzweig}
\author{Anuoluwapo Bolarinwa}
\author{Alexandre Gramfort}
\author{Michael I Mandel}
\affil{Reality Labs, Meta}

\begin{document}

\maketitle

\setcounter{footnote}{0}

\begin{abstract}
Surface electromyography (sEMG) non-invasively measures signals generated by muscle activity with sufficient sensitivity to detect individual spinal neurons and richness to identify dozens of gestures and their nuances. Wearable wrist-based sEMG sensors have the potential to offer low friction, subtle, information rich, always available human-computer inputs. To this end, we introduce \emgtwoqwerty, a large-scale dataset of non-invasive electromyographic signals recorded at the wrists while touch typing on a QWERTY keyboard, together with ground-truth annotations and reproducible baselines\footnote{\url{https://github.com/facebookresearch/emg2qwerty}}. With 1,135 sessions spanning 108 users and 346 hours of recording, this is the largest such public dataset to date. These data demonstrate non-trivial, but well defined hierarchical relationships both in terms of the generative process, from neurons to muscles and muscle combinations, as well as in terms of domain shift across users and user sessions. Applying standard modeling techniques from the closely related field of Automatic Speech Recognition (ASR), we show strong baseline performance on predicting key-presses using sEMG signals alone. We believe the richness of this task and dataset will facilitate progress in several problems of interest to both the machine learning and neuroscientific communities.
\end{abstract}

\section{Introduction}
\label{sec:intro}


The bandwidth of communication from computers to humans has increased dramatically over the past several decades through the development of high fidelity visual and auditory displays~\citep{KOCH20061428}. The bandwidth from humans to computers, however, has remained severely rate-limited by keyboard, mouse, and touch screen control for the vast majority of applications---modalities that have changed little in the past 50 years. One potential approach to increasing human-to-machine bandwidth is to leverage the high dimensional output of the human peripheral motor system. Ultimately, it is this system that evolved to perform all human actions on the world. 

Non-invasive interfaces based on electromyographic signals (sEMG) capturing neuro-muscular activity at the wrist~\citep{ctrl-labs2024}, have the potential to achieve higher bandwidth human-to-machine input by measuring not just the activity of individual muscles, but also the individual motor neurons that control them. With each muscle composed of hundreds of motor units~\citep{floeter_karpati_2010, enoka2013motor}, this expansion has the potential to unlock orders of magnitude higher bandwidth if subjects can learn to control them individually~\citep{harrison1962identification, Basmajian1963}. In comparison, methods that capture brain activity non-invasively such as fMRI or EEG do not offer the resolution of individual action potentials, and are often cumbersome for broader applicability beyond clinical or in-lab settings.

More generally, the transduction of neural signals (peripheral or central) to language (text or speech) has the potential for broad applications, as evidenced by a number of recent works using intracranial recordings~\citep{fan2023plugandplay,Metzger-etal:2023,willett2023high} or non-invasive EEG or MEG signals~\citep{duan2023dewave,defossez-etal:2023}. The rapid adoption of mobile computing has been at the cost of reduced human-computer bandwidth, via thumb-typed and swipe-based text entry, compared to the era of desktop computing. In existing augmented and virtual reality (AR/VR) environments, text is input through cumbersome point-and-click typing or speech transcription. While speech-to-text systems have a lower barrier for adoption, they are largely only suitable for tasks that can be formulated as dictation or conversation, whereas neuromotor interfaces can additionally be utilized for a broad range of realtime motor tasks like robotic control. Furthermore, the wider adoption of speech interfaces is limited by social privacy implications, in particular in public and semi-public (e.g., open-plan office) settings. For these reasons, a high bandwidth and private text input system via sEMG, such as typing without a keyboard with wrists resting on laps or a flat surface, can be highly useful for for AR/VR environments.

While wrist-based sEMG is a flexible and practical human-computer input modality, progress has been challenging due to the cumbersome nature of data collection, and the lack of large-scale datasets with well-defined tasks and measurable benchmarks. To facilitate scientific progress, and in anticipation of the potentially wide adoption of sEMG measurement hardware that can provide both user convenience and high signal quality, we introduce a large dataset for the task of predicting key-presses while touch typing on a keyboard using sEMG activity alone. Transcribing typing from sEMG is analogous to Automatic Speech Recognition (ASR) in that a fixed-rate sequence of continuous high dimensional features is transduced into a sequence of characters or words. It has a complicated, but well defined input-output relationship that is amenable to modeling while remaining highly non-trivial.

As a machine learning problem, the task of transcribing key-presses from sEMG is interesting and challenging for several reasons. Like ML systems based on EEG, there is a great deal of domain shift across sessions~\citep{Kobler-eta:2022,bakas2023latent,gnassounou2023convolution}. We will refer to a session as one episode of one user donning and doffing the band. This domain shift has at least three causes: cross-user variation due to differences in anatomy and physiology, cross-session variation due to differences in band placement relative to the anatomy, and cross-user behavioral differences due to different typing strategies (e.g., finger-to-key mapping, force level, response to keyboard properties like key travel, etc.). The benchmarks included with this dataset are designed to explicitly characterize the ability of models to generalize across these various forms of domain shift. To give a sense of the magnitude of these domain shifts, the unpersonalized performance on a new user in our benchmark is over 50\% character error rate (Section~\ref{sec:experimental_results}), indicating that the model cannot successfully transcribe the majority of keystrokes without some labeled data from the user.

A second interesting aspect of the problem is that, while similar in structure to speech signals, the process through which sEMG is generated is quite different. In particular, a good approximation to the speech production process is the application of a time-varying filter to a time-varying source~\citep{Fant1971}. sEMG, on the other hand, is generated through a purely additive combination of muscles in a process described in Section~\ref{sec:emg_background}. Furthermore, the speech signal itself has developed so as to be directly interpreted by a wide range of listeners, whereas sEMG is one step removed from direct interpretability. Specifically in the case of typing, it is the key-press itself that is the target output, and there are fewer constraints on consistency across individuals in how this is achieved.

A third interesting aspect, and contrast to ASR, is the difference between spoken language and typing. In particular, typing occurs directly at the character level, whereas speech consists of a sequence of phonemes, which are mapped via a complex and heterogeneous process to written characters (especially for English). This difference is most apparent in the use of the backspace key in typing, which language models operating on sEMG-to-text predictions must account for, unlike in spoken language. Furthermore, while the transcription of spoken language can result in homophones---words that are pronounced the same but spelled differently---no such ambiguity is present in typed text. Finally, it is possible to type character strings that are difficult or impossible to speak and are certainly outside of a given lexicon. This is particularly relevant when it comes to entering passwords or URLs, but more generally is applicable when using keyboards for tasks other than typing full sentences.

In this paper, we introduce \emgtwoqwerty, a dataset of sEMG signals recorded at the wrists while typing on a QWERTY keyboard. To our knowledge, this is the largest such public sEMG dataset to date with recordings from 108 users and 1,135 sessions spanning 346 hours, and with precise ground-truth annotations. We describe baseline models leveraging standard ASR components and methodologies, benchmarks to evaluate zero-shot generalization to unseen population and data-efficient personalization, and reproducible baseline experimental results against these benchmarks. We conclude with open problems and future directions.

\section{Background on sEMG}
\label{sec:emg_background}

\begin{figure}
\centering
\begin{tabular}{cc}
\includegraphics[width=0.75\columnwidth, keepaspectratio]{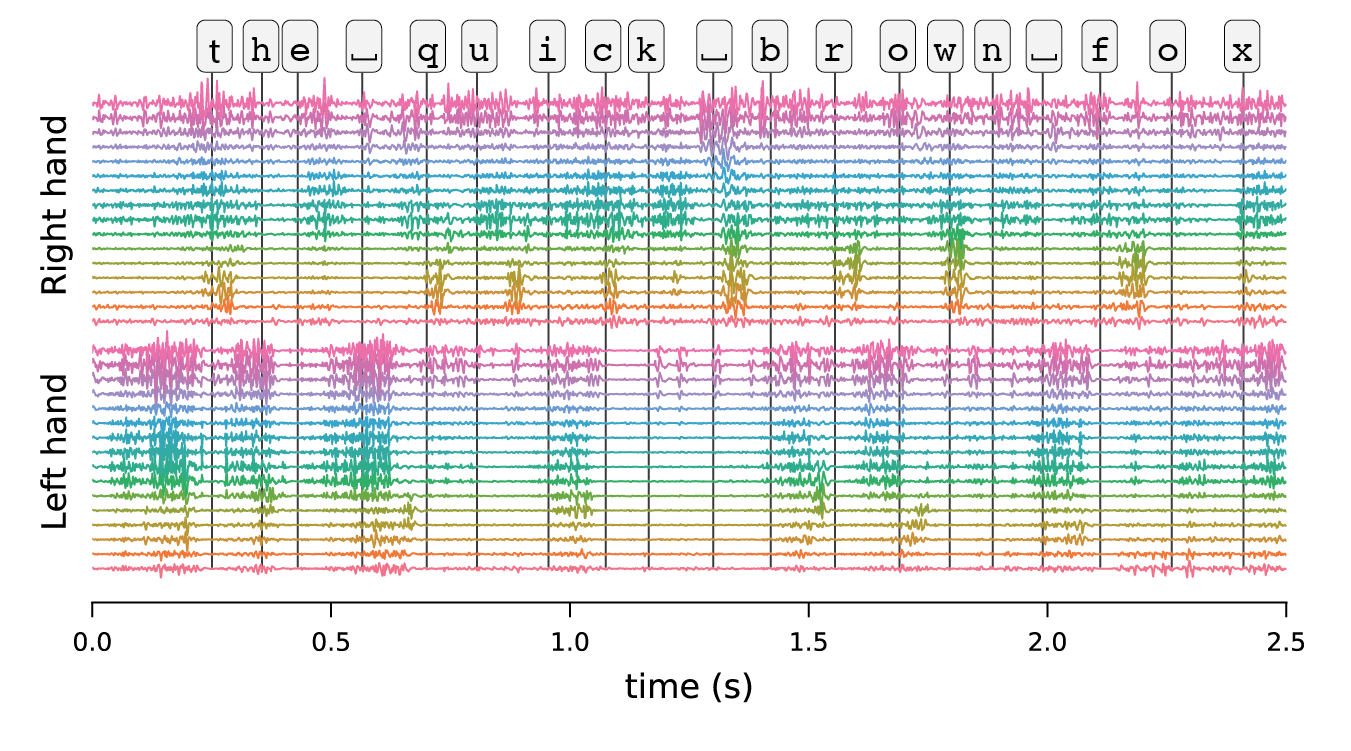}  &
\includegraphics[width=0.25\columnwidth, keepaspectratio] {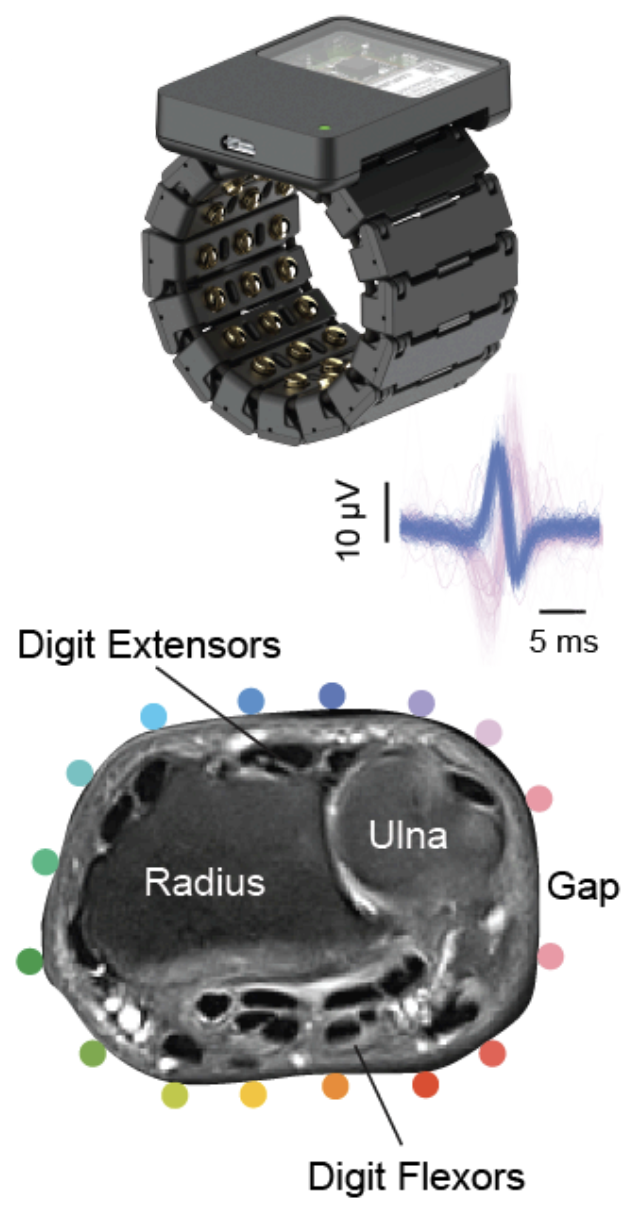}
\end{tabular}
\caption{\textbf{Left:} An example surface electromyographic (sEMG) recording for the prompt ``the quick brown fox'' showing 32-channel sEMG signals from left and right wristbands, along with key-press times. Vertical lines indicate keystroke onset. The signal from each electrode channel is high-pass filtered. \textbf{Right:} The sEMG research device (sEMG-RD) used for data collection together with a schematic denoting the electrode placement around the wrist circumference. The left and right wristbands are worn such that one is a mirror of the other, and therefore the positioning of the electrodes around the wrist physiology remains the same, albeit with a reversed electrode polarity with respect to the wrist.}
\label{fig:exampleDataSegment}
\end{figure}

Surface electromyography (sEMG) is the measurement at the skin surface of electrical potentials generated by muscles~\citep{merletti2016surface}. It has the ability to detect the activity caused by individual motor neurons while being non-invasive. Specifically, a single spinal motor neuron, with cell body in the spinal cord, projects a long axon to many fibers of a single muscle. Each muscle fiber is innervated by only one motor neuron. When that neuron fires, it triggers all of the muscle fibers that it innervates to contract, and in the process they generate a large electrical pulse, in effect amplifying the pulse from the neuron. It is this electrical signal from the muscle fibers that sEMG sensors on the skin can detect.

A single motor neuron and the muscle fibers that it innervates are known together as a \emph{motor unit}. Muscles vary widely in terms of how many motor units they contain as well as how many muscle fibers each motor unit contains. For the hand and forearm, muscles contain on the order of hundreds of motor units, each of which contains hundreds of muscle fibers \citep{floeter_karpati_2010, kandel2013principles}. One interesting property of the sEMG signal relevant to brain-computer interfaces is that, because it is generated during muscle fiber stimulation, it typically precedes the onset of the corresponding motion by tens of milliseconds \citep[e.g.,][]{TrigiliEtAl2019}. This can be seen in Figure~\ref{fig:exampleDataSegment} where the left hand's sEMG shows a strong activation before the ``t'' key is pressed and the right hand before the subsequent ``h''. This property means that sEMG-based interfaces have the potential to detect activity with \emph{negative} latency, i.e., before the corresponding physical gesture occurs.

With regards to machine learning modeling, the motor system is organized in a very structured way. Typically, within a given muscle, there is a more or less fixed order in which motor units are \emph{recruited} as a function of force generated by that muscle~\citep{henneman1957, deluca2012hierarchical}. Thus there is a motor unit that is generally recruited first, which is activated before the motor unit that is generally recruited second, etc. While this was traditionally shown to be true for isometric contractions (i.e., with joints held at fixed angles, and therefore muscles at fixed lengths), there are some recent~\citep{Formento2021noninvasive, marshall2022flexible, hug2023common} and less recent works~\citep{harrison1962identification, Basmajian1963} showing that this ordering may depend on additional factors such as muscle length or target motion.


\section{Related Work}
\label{sec:related_work}


\begin{table}[t]
    \caption{Comparison with prior electromyographic datasets}
    \label{table:related_datasets}
    \centering
    
    \begin{tabularx}{\textwidth}{l X l X X X}
        \toprule
        Dataset & Hardware grade & Application & Recording location & Subject count & Multiple sessions/subject \\
        \midrule
        Amma et al. 2015      & Clinical & HCI                    & Forearm            & 5               & Yes \\
        Du et al. 2017        & Clinical & HCI                    & Forearm            & 23              & Yes \\
        Malesevic et al. 2021 & Clinical & Neuroprosthetics       & Forearm            & 20              & No           \\
        Jiang et al. 2021     & Clinical & HCI, Neuroprosthetics  & Wrist              & 20              & Yes           \\
        Ozdemir et al. 2022   & Clinical & Neuroprosthetics       & Forearm            & 40              & No           \\
        Kueper et al. 2024    & Clinical & Neuroprosthetics       & Forearm            & 8               & Yes           \\
        Palermo et al. 2017   & Clinical & Neuroprosthetics       & Forearm            & 10              & Yes           \\
        Atzori et al. 2012    & Consumer & Neuroprosthetics & Forearm, Wrist     & 27              & No           \\
        Pizolato et al. 2017  & Consumer & Neuropresthetics & Forearm, Wrist     & 78              & No           \\
        Lobov et al. 2018     & Consumer & Neuropresthetics & Forearm            & 37              & No           \\
        \emgtwoqwerty         & Consumer & HCI                    & Wrist              & 108             & Yes           \\
        \bottomrule
    \end{tabularx}
\end{table}

The direct mapping of neuromotor activity to text enables high bandwidth human-computer interfaces for regularly abled people, but brain-computer interfaces could be applicable even for those without full use of their limbs. \citet{willetEtAl2021} demonstrated an intracortical brain-computer interface that allowed a single subject with a paralyzed hand to generate text at a rate of 90 characters per minute through imagined writing motions. Speech decoding from cerebral cortical activity has been demonstrated in paralyzed participants~\citep{Moses2021, willett2023high}, where latest results achieve 62 words per minute on large vocabularies while decoding sentences in real time with 25.6\% word error rate. Modeling methodologies that prove successful on the \emgtwoqwerty benchmark could be adapted with some care for use in such brain-computer interfaces.


While there are several public sEMG datasets, \emgtwoqwerty is unique in its hardware properties, scale, and the type of participant activity. In terms of hardware, existing sEMG datasets use either clinical-grade high-density electrode arrays and amplifiers~\citep{AmmaEtAl2015, Du_Sensors_2017, MalesevicEtAl2021, JiangEtAl2021, ozdemir2022dataset, kueper2024eeg}, or consumer-grade hardware that only records coarse sEMG energy at low sampling rates~\citep{AtzoriEtAl2012, PizzolatoEtAl2017, LobovEtAl2018}. The clinical-grade setup records hundreds of channels of sEMG with high sampling rate and bit depth, but requires careful setup including shaving and abrading the skin, applying conductive gel, and taping the electrode array(s) in place for the duration of the recording. The now discontinued consumer-grade Myo armband by Thalmic Labs can be worn without any preparation, but only measures 8 channels of sEMG and has a low sampling rate (200~Hz) after the signal has been rectified and smoothed. On the other hand, \emgtwoqwerty uses a new research-grade dry electrode device, sEMG-RD~\citep{ctrl-labs2024}, that offers consumer-grade practicality while providing signal quality approaching that of the clinical-grade setup.

In terms of scale, \emgtwoqwerty is not only among the largest in raw hours of data, but also in the number of subjects and number of sessions per subject. Many of the existing datasets only include a single recording session per subject \citep{AtzoriEtAl2012, AtzoriEtAl2014, PizzolatoEtAl2017, LobovEtAl2018, MalesevicEtAl2020, MalesevicEtAl2021, ozdemir2022dataset}, although a small number include more. In particular, \citet{PalermoEtAl2017} include 10 subjects with 10 sessions each, \citet{AmmaEtAl2015} include 5 subjects with 5 sessions each, \citet{Du_Sensors_2017} include 23 subjects with 3 sessions for 10 of them, \citet{JiangEtAl2021} include 20 subjects with 2 sessions each, and \citet{kueper2024eeg} include 8 subjects with 10 sessions each. In comparison, \emgtwoqwerty is over an order of magnitude larger, with 108 subjects and an average of 10 sessions per subject, making it feasible to evaluate the ability of models to generalize both across sessions and across subjects. This focus on broader generalization makes \emgtwoqwerty more challenging than the existing benchmarks.

In terms of activity, \emgtwoqwerty focuses on the natural behavior of typing, while existing datasets focus for the most part on static hand poses. \citet{AtzoriEtAl2012} introduced a set of 50 hand gestures, many of which may be considered ``abstract'' in that they are not natural movements that a subject would spontaneously make in everyday life, such as the flexion or extension of individual fingers or pairs of fingers. A subset of 23 of the gestures does however focus on grasping and functional movements. Each gesture is held for 5 seconds as a static posture. Many subsequent datasets have utilized the same or similar gestural vocabularies \citep{AtzoriEtAl2014, PizzolatoEtAl2017, AmmaEtAl2015, Du_Sensors_2017, LobovEtAl2018, MalesevicEtAl2021, JiangEtAl2021, ozdemir2022dataset, kueper2024eeg}. One exception to the use of static poses is \citet{MalesevicEtAl2020}, who prompt users to follow temporally evolving cues. However, they are recorded intra-muscularly using inserted electrodes rather than non-invasively on the surface. In contrast, we present non-invasive sEMG recordings of the natural, rapidly time-varying task of typing on a keyboard. Keystroke events occur much more rapidly than held poses, and the mean keystroke rate in our dataset is 4.4 characters per second (Table~\ref{table:dataset_details}). Moreover, subjects are much more familiar with typing than with performing abstract movements of their individual hand joints. Typing style does, however, vary considerably across individuals, especially those who are not fluent touch typists, adding to the challenge of generalization across subjects.

\section{Dataset and Benchmark}
\label{sec:data}

\begin{table}[t]
    \caption{\emgtwoqwerty dataset statistics}
    \label{table:dataset_details}
    \centering
    \begin{tabular}{ l  r @{~} l } 
        \toprule
        Total subjects & 108 & \\
        Total sessions & 1,135 & \\
        Avg sessions per subject & 10 & \\
        Max sessions per subject & 18 & \\
        Min sessions per subject & 1 & \\
        \midrule
        Total duration & 346.4 & hours \\
        Avg duration per subject & 3.2 & hours \\
        Max duration per subject & 6.5 & hours \\
        Min duration per subject & 15.3 & minutes \\
        \midrule
        Avg duration per session & 18.0 & minutes \\
        Max duration per session & 47.5 & minutes \\
        Min duration per session & 9.5 & minutes \\
        \midrule
        Avg typing rate per subject & 265 & keys/min \\
        Max typing rate per subject & 439 & keys/min \\
        Min typing rate per subject & 130 & keys/min \\
        \midrule
        Total keystrokes & 5,262,671 & \\
        \bottomrule
    \end{tabular}
\end{table}

\paragraph{sEMG-RD Hardware}

All data were recorded using the sEMG research device (sEMG-RD) described in \citet{ctrl-labs2024} and visualized in Figure~\ref{fig:exampleDataSegment}. Each sEMG-RD has 16 differential electrode pairs utilizing dry gold-plated electrodes. Signals are sampled at 2~kHz with a bit depth of 12 bits and a maximum signal amplitude of 6.6~mV. Measurements are bandpass filtered with -3~dB cutoffs at 20~Hz and 850~Hz before digitization. Data are digitized on the sEMG-RD and streamed via Bluetooth to the laptop that the subject is simultaneously typing on. Identical devices are worn on the left and right wrists, with the same electrode indices aligning with the same anatomical features, but the polarity of the differential sensing reversed. 

\paragraph{Data Collection Setup}

Data collection participants were initially screened for their touch typing ability via self-reporting. Those reported as being able to type without looking at the keyboard and achieve the correct finger to key mapping at least about 90\% of the time subsequently took part in a typing test, and only those meeting adequate typing speed and accuracy levels were included in the study. In the spirit of making the dataset realistic for broader usage where the majority of real world touch typists are not perfect in their finger to key mapping, we do not enforce a rigid constraint beyond the aforementioned screening process.

Participants partake in a few sessions of typing prompted text after donning two sEMG-RDs, one per wrist. During each session, the participant is prompted with a sequence of text that they type on an Apple Magic Keyboard (US English). The text prompts consist of words sampled randomly from a dictionary as well as sentences sampled from English Wikipedia (after filtering out offensive terms), and are processed to only contain lower-case and basic punctuation characters. A keylogger records the typed characters, together with key-down and key-up timestamps, as ground-truth. Participants are allowed to freely use the backspace key to correct for typos. Session duration varied from 9.5 to 47.5 minutes depending on the typing speed of the participant (Table~\ref{table:dataset_details}). Between sessions, the bands were doffed and donned to include realistic variability in electrode placement and therefore the recorded signals. Figure~\ref{fig:exampleDataSegment} visualizes an example sEMG recording. 

This study was approved by an Institutional Review Board (IRB), participation was on a purely voluntary basis, participants provided informed consent to include their data, and the data have been de-identified by removing any personally identifiable information. Subjects were allowed to withdraw from the study during their participation.

\paragraph{Dataset Details}

Recorded sessions undergo minimal post-processing---basic signal quality checks, high-pass filtering with 40~Hz cutoff, and an algorithm to correct for clock drift and synchronize timestamps across the sEMG devices and the host laptop. The signals from the left and right bands are temporally aligned by interpolating to the nearest sample (i.e., 0.5~ms). The entire dataset consists of 1,135 session files from 108 participants and 346 hours of recording (Table~\ref{table:dataset_details}). Each session file follows a simple HDF5 format comprising the aligned left and right sEMG signal timeseries spanning the duration of the recording, ground-truth keylogger sequence, timestamps corresponding to sEMG and the keystrokes, and additional metadata. To facilitate usage of the data in the neuroscience community, a script is provided to convert the data to BIDS format~\citep{pernet-etal:2019,Poldrack-etal:2024}.

\paragraph{Benchmark Setup}

\begin{figure}[t!]
    \centering
    \includegraphics[width=1\textwidth, keepaspectratio]{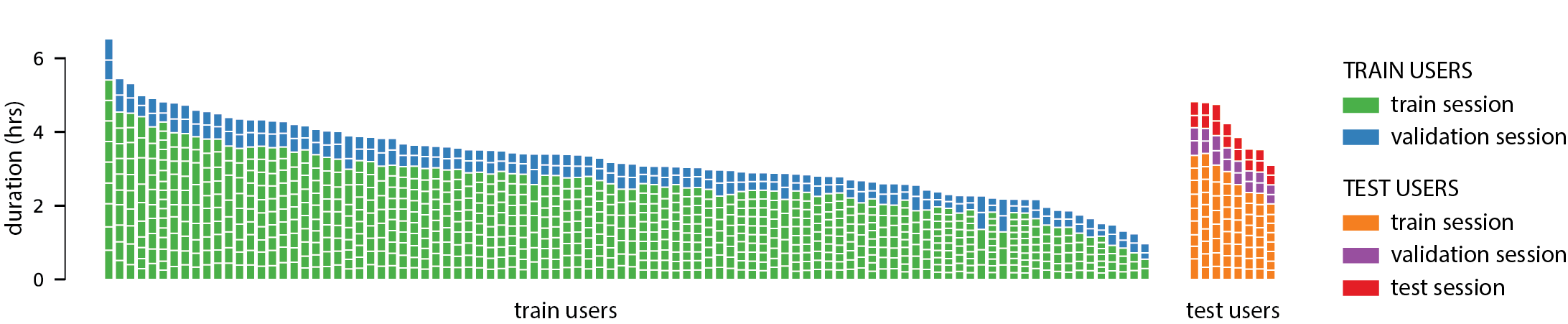}
    \caption{Visualization of \emgtwoqwerty dataset splits. Each column represents a user. Each block represents a session, with the vertical extent of each block indicating its duration. Train users correspond to the data used for training a single generic model. A personalized model is produced for each of the 8 test users. Sessions used for training, validation and testing are color-coded.}
    \label{fig:benchmark_setup}
\end{figure}

The key difficulties in building models based on sEMG at scale are of generalization to unseen population and data-efficiency of personalization. Differences in human physiology, high variance in typing behavior, and complexity of data acquisition all contribute to the problem. With this mind, we define our dataset splits as follows: we sample 8 test users out of the 108 participants to be held out for personalization. For each of these 8 users, we hold out 2 sessions each for validation and testing, and the remaining sessions are used to train per-user personalized models. The sessions from the remaining 100 users are used to train a generic user model which can then be finetuned and personalized to each of the 8 test users. Figure~\ref{fig:benchmark_setup} visualizes the data splits.

\paragraph{Metric}
We measure the predictive power of the models in terms of \textit{Character Error Rate} (CER), which we define as the \textit{Levenshtein} edit-distance between the predicted and the ground-truth sequences of keystrokes, normalized by the length of the ground-truth sequence. The CER of the generic model on the test sessions from each of the 8 held-out users measures generalization to unseen population. The same evaluation of the personalized models measures the data-efficiency of personalization.


\section{Baselines}
\label{sec:baselines}

\subsection{Baseline Model}
\label{sec:baseline_model}

The similarity between the \emgtwoqwerty task and that of recognizing speech from audio waveforms allows us to borrow from the field of ASR in our modeling approach. Both tasks map continuous waveform signals (1D for ASR, 32D for \emgtwoqwerty) at a fixed sample rate, to a sequence of tokens (phonemes or words for ASR, characters for \emgtwoqwerty). The components of our baseline model spanning preprocessing and feature extraction, data augmentation, model architecture, loss function, language model and decoder are largely applications of ASR methodologies.

\paragraph{Feature Extraction}

In ASR, models commonly use log mel-scale filter banks for features, although fully end-to-end models trained from the raw waveform have been explored \citep{palaz2013estimating, DBLP:conf/interspeech/TuskeGSN14, RonWeissICAASP2015, zeghidour2018end, baevski2020wav2vec}. The mel scale is adapted from human auditory perception, and thus is not necessarily appropriate for the spectral characteristics of sEMG signals. We use analogous spectral features on a different log frequency scale with appropriate cutoffs. Spectral features outperformed rectification of the time domain signal, which is a standard preprocessing method for sEMG~\citep{halliday2010need}. As a way of normalizing the spectral features, we add a 2D batch normalization step as the first layer of our network that computes per-channel spectrogram statistics.

\paragraph{Data Augmentation}

SpecAugment~\citep{park2019specaugment} is a simple but effective data augmentation strategy in ASR. It applies time- and frequency-aligned masks to spectral features during training. We find modest but consistent improvements with SpecAugment in our baseline model. Additionally, we include two other forms of data augmentation that are specific to our use case: 1) \emph{rotation augmentation} rotates (permutes) the electrodes by $-1$, $0$, or $+1$ positions sampled uniformly, meaning, the electrode channels are all shifted one position to the left, remain unshifted, or are shifted one position to the right respectively; 2) \emph{temporal alignment jitter} randomly jitters the alignment between left and right sEMG timeseries in the raw signal space by an offset value sampled uniformly in the range of $0$ to $120$ samples (60~ms for 2~kHz signal).

\paragraph{Model Architecture}

We adopt Time Depth Separable ConvNets (TDS) developed by \citet{hannunEtAl2019} for the ASR domain. The parameter-efficiency of TDS convolutions allows for wider receptive fields which have proven important in \emgtwoqwerty modeling. While the sEMG activity profile corresponding to a single key-press is fairly short, ``co-articulation'' activity is dominant in the signal. Specifically, the sEMG of a keystroke is affected by the characters typed immediately before and after it due to various preparatory behaviors. We thus find it effective to use a model with a receptive field long enough to capture sEMG activity related to bigrams or trigrams, especially for fast typists, but not so long as to essentially learn an implicit language model. The TDS architecture allows parameter-efficient control of this trade-off, and our baseline model uses a receptive field of 1 second. Alternative architectures such as convolutional ResNets, RNNs and transformers have all been effective in ASR \citep{synnaeve2019end}, though we do not explore them here. Additionally, our model architecture includes two ``Rotation-Invariance'' modules, one for each band. Each of these is composed of a linear layer and a ReLU activation, which get applied to electrode channel shifts of $-1$, $0$, and $+1$ positions, and their outputs averaged over. We concatenate the outputs of the Rotation-Invariance module corresponding to each band prior to being fed into the TDS network. The Rotation-Invariance modules, together with rotation augmentation, aim to improve the model's generalization across a user's sessions, since doffing and donning each band between sessions can result in electrode shifts corresponding to small rotations of the band's placement on the wrist.

\paragraph{Loss Function}

In ASR, the alignment problem of identifying precise timings of output tokens can be circumvented using losses such as Connectionist Temporal Classification (CTC)~\citep{gravesEtAl2006}, sequence-to-sequence (Seq2Seq)~\citep{chorowski2015attentionbased, chan2016listen}, or RNN-transducer (RNN-T)~\citep{graves2012sequence}. \emgtwoqwerty includes precise key-press and key-release event timings, allowing for the use of a cross-entropy classification loss averaged over all time points in the output sequence. Nevertheless, we empirically find the best performance with CTC loss which we therefore use in our baseline.

\paragraph{Language Model and Decoder}

At test time, we apply a simple, lexicon-free, character-level n-gram language model (LM) to the model predictions. In our experiments, we use a 6-gram modified Kneser-Ney LM~\citep{heafield-etal-2013-scalable} generated from WikiText-103 raw dataset~\citep{DBLP:journals/corr/MerityXBS16}, and built using the KenLM package~\citep{heafield-2011-kenlm}. The LM is integrated with the CTC logits using an efficient first-pass beam-search decoder similar to~\citet{Pratap_2019}. As noted earlier, the ability to modify history using the backspace key makes this a more complex task compared to ASR decoding. Therefore, we implement a modified backspace-aware beam-search decoder that safely updates LM states, which we include in our open-source repository.

\subsection{Training Setup}
\label{sec:training_setup}

With the data splits described in Section~\ref{sec:data}, we train one generic user model with sessions from 100 users, as well as 8 independent personalized models for each of the held-out test users. For the generic model, 2 sessions from each of the 100 users are used for validation. This ensures that the test users do not bear any influence on the hyperparameter choices of the generic user model. The personalized models for test users are trained using the splits defined in Section~\ref{sec:data}, and are initialized either with random weights or with the weights of the generic model.

All training runs use the architecture described in Section~\ref{sec:baseline_model}. Training and validation operate over batches of contiguous 4~second windows, whereas at test time, we feed entire sessions at once without batching to avoid padding effects on test scores. We train using windows asymmetrically padded as in \citet{pratap2020scaling}, with 900~ms past and 100~ms future contexts, to minimize the dependency on future inputs and facilitate streaming applications. The input to the network are 33-dimensional log-scaled spectral features for each of the 32 electrode channels spanning the left and right bands, computed every 8~ms with a 32~ms window. We apply SpecAugment style time masking (maximum 3 masks of segments of length up to 200~ms) and frequency masking (maximum 2 masks of up to 4 contiguous frequency bins) on the log spectrogram features for each of the 32 channels i.i.d. We also apply band rotation and temporal alignment jitter augmentations described in Section~\ref{sec:baseline_model}. The $33 \times 16 = 528$ spectral features per band are input into their respective Rotation-Invariance modules following batch normalization. Each Rotation-Invariance module outputs 384-dimensional features which are then concatenated and fed into a sequence of 4 TDS convolutional blocks, each with a kernel width of 32 temporal samples. The network, in conjunction with the spectrogram features, has a total receptive field of 1~second (125 samples).

The generic model was trained on 8 A10 GPUs and each of the personalized models on a single A10 GPU, using a batch size of 32 per GPU, and optimized with Adam~\citep{kingma2017adam} for a total of 150 epochs. The learning rate linearly ramps every epoch from 1e-8 to 1e-3 for the first 10 epochs, and then follows a cosine annealing schedule~\citep{cosine_annealing} towards a minimum of 1e-6 for the remaining epochs. Both the generic and personalized models share the same set of hyperparameters, which were initialized to reasonable values without investing in a thorough sweep as that is not the focus of the paper. Our code uses PyTorch \citep{NEURIPS2019_9015} and PyTorch Lightning \citep{falcon2019pytorch} for training, and Hydra \citep{Yadan2019Hydra} for configuration.

\subsection{Experimental Results}
\label{sec:experimental_results}

\begin{table}[t]
    \caption{A comparison of test subject performance across model benchmarks. Mean and standard deviation aggregates of character error rates (CER) across test subjects are reported. Lower is better. The reported test CER improvements arising out of personalization as well as the inclusion of the language model (LM), all have $p < .005$.}
    \label{table:experimental_results}
    \centering
    \begin{tabularx}{\textwidth}{ l r r r r } 
        \toprule
        & \multicolumn{2}{c}{No LM} & \multicolumn{2}{c}{6-gram char-LM} \\
        \cmidrule(lr){2-3} \cmidrule(lr){4-5}
        Model benchmark & Val CER & Test CER & Val CER & Test CER \\ 
        \midrule
        Generic (no personalization) & $55.57 \pm 4.40$ & $55.38 \pm 4.10$ & $52.10 \pm 5.54$ & $51.78 \pm 4.61$ \\
        \midrule
        Personalized (random-init) & $15.65 \pm 5.95$ & $15.38 \pm 5.88$ & $11.03 \pm 4.45$ & $9.55 \pm 5.16$ \\
        \midrule
        Personalized (finetuned) & $11.39 \pm 4.28$ & $11.28 \pm 4.45$ & $8.31 \pm 3.19$ & $6.95 \pm 3.61$ \\
        \bottomrule
    \end{tabularx}
\end{table}

For each of the 8 held-out users, we evaluate performance on their validation and test sessions with (1) the generic model, (2) respective personalized models initialized with random weights, (3) respective personalized models initialized with the generic model weights. Table~\ref{table:experimental_results} reports the mean and standard deviation of the character error rates (CER) aggregated over the 8 test users for each of the three scenarios, without and with the usage of the language model (LM) described in Section~\ref{sec:baseline_model}.

Not surprisingly, the approach of finetuning the generic model with user-specific sessions to obtain personalized models performs the best by achieving a mean test CER of 6.95\% (with LM). The best performing user achieves a test CER as low as 3.16\%. We note that, in practice, models tend to become usable at a CER of approximately 10\% or less. To further ground this, \citet{perola_mobilekeyboard} find an uncorrected error rate (percentage of errors remaining after user editing) of 2.3\% from a large-scale study of typing on a mobile keyboard that could include autocorrect and other assistance.

``Personalized (finetuned)'' in Table~\ref{table:experimental_results} outperforming ``Personalized (random-init)'' across all metrics demonstrates that generalizable representations of sEMG can indeed be learned across users, despite variations in sensor placement, anatomy, physiology, and behavior. When evaluated directly against unseen subjects though, the generic model is simply unusable with a CER over 50\%. This can be attributed to the scale of the generic model in terms of the number of users it has been trained with. \citet{ctrl-labs2024} demonstrate that performant out-of-the-box generalization of sEMG models across people can indeed be achieved with an order of magnitude more training users, albeit for different tasks. Still, it is exciting to see generalization emerge even at the scale of 100 users, and motivates research into data-efficient strategies to improve generalization further to alleviate the need for cumbersome and expensive large-scale supervised data collection.



\section{Limitations}
\label{sec:limitations}

The biggest limitation is that \emgtwoqwerty models require the skill of touch typing on a QWERTY keyboard in addition to being limited only to English. Although this dataset is meant to be a starting point, these imply that the generated models would only be applicable to a subset of the society.

Additionally, the dataset was collected using a physical keyboard on a desk, whereas real-world use cases would be aimed at removing physical constraints to enable seamless text entry such as typing while on the move or with hands on the lap. Moreover, the force generated while typing on a physical keyboard, and thus the amplitude of the detected sEMG, might be different compared to keyboard-free sEMG-based typing, leading to a domain mismatch.

During deployment, the model would need to run as close to the source of the signal as possible, preferably on the wristband itself. This is ideal for reasons around latency, user privacy, and avoidance of Bluetooth contention or interference arising from streaming sEMG off the wristband. While compute and battery constraints on the wristband might pose a challenge, the advent of low-cost edge inference for machine learning makes this a practical solution.

Finally, the lack of broader access to the proprietary research-grade sEMG hardware might be limiting in some ways such as not being able to perform human-in-the-loop testing of models.

Further ethical and societal implications are discussed in Appendix~\ref{app:ethics}.

\section{Conclusion and Future Directions}
\label{sec:conclusion}

sEMG-based typing interfaces operating at a rate of hundreds of characters per minute, demonstrate the feasibility of practical and scalable neuromotor interfaces, and set the stage for increases in human-to-computer bandwidth. These could also be the starting point for highly personalizable neuromotor interfaces that adapt to minimize the amount of physical effort necessary, while simultaneously customizing the action-to-intent mapping based on task, context, and stylistic preference.

There is a lot of excitement about the potential of practical brain-computer or peripheral neuromotor interfaces. Yet, to date there has been no satisfactory public dataset with the scale adequate to build broadly applicable systems leveraging the strides in the field of machine learning. We believe \emgtwoqwerty is an early step towards addressing this gap. In contrast to prior sEMG systems that focus on neuroprosthetics or clinical settings, our focus is on a practical high bandwidth input interface for AR/VR that can work across the population. We demonstrate that standard paradigms and off-the-shelf components from the closely related field of ASR, facilitate the creation of models that enter the realm of usability. Our benchmarks empirically quantify the difficulty of the task arising out of physiological and behavioral domain shift, and offer a yard stick to measure progress.

For the machine learning community, we hope \emgtwoqwerty will be a useful benchmark for existing research problems. Advances in areas such as domain adaptation, self-supervision, end-to-end sequence learning and differentiable language models will serve to benefit this task. For the neuroscience community, we believe access to a large sEMG dataset opens up new research avenues. Advances in white-box feature extraction methods such as unsupervised spike detection and sorting could enable the construction of models that are robust to variability in electrode placement and physiology.


\section*{Acknowledgments}
We thank Joseph Zhong for implementing the data collection protocol, Gabriel Synnaeve and Awni Hannun for sharing their ASR wisdom, Rudi Chiarito, Sam Russell and Krunal Naik for engineering support, Dan Wetmore, Ron Weiss and Simone Totaro for their feedback on the paper, Vittorio Caggiano and Nafissa Yakubova for helpful discussions, Patrick Kaifosh and Thomas R. Reardon for their vision and sponsorship, and the entire CTRL-labs team whose efforts this work builds upon.


\bibliographystyle{abbrvnat}
\bibliography{main.bib}


\section*{Checklist}

\begin{enumerate}

\item For all authors...
\begin{enumerate}
  \item Do the main claims made in the abstract and introduction accurately reflect the paper's contributions and scope?
    \answerYes{}
  \item Did you describe the limitations of your work?
    \answerYes{See Section~\ref{sec:limitations}.}
  \item Did you discuss any potential negative societal impacts of your work?
    \answerYes{{See Section~\ref{sec:limitations} and Appendix~\ref{app:ethics}.}}
  \item Have you read the ethics review guidelines and ensured that your paper conforms to them?
    \answerYes{}
\end{enumerate}

\item If you are including theoretical results...
\begin{enumerate}
  \item Did you state the full set of assumptions of all theoretical results?
    \answerNA{}
	\item Did you include complete proofs of all theoretical results?
    \answerNA{}
\end{enumerate}

\item If you ran experiments (e.g. for benchmarks)...
\begin{enumerate}
  \item Did you include the code, data, and instructions needed to reproduce the main experimental results (either in the supplemental material or as a URL)?
    \answerYes{The code and data, with instructions to reproduce experimental results, are publicly accessible at \href{https://github.com/facebookresearch/emg2qwerty}{github.com/facebookresearch/emg2qwerty}.}
  \item Did you specify all the training details (e.g., data splits, hyperparameters, how they were chosen)?
    \answerYes{See Sections \ref{sec:data} and \ref{sec:training_setup}.}
	\item Did you report error bars (e.g., with respect to the random seed after running experiments multiple times)?
    \answerYes{Table~\ref{table:experimental_results} reports error bars obtained over experiments run with multiple randomly sampled users.}
	\item Did you include the total amount of compute and the type of resources used (e.g., type of GPUs, internal cluster, or cloud provider)?
    \answerYes{See Section~\ref{sec:training_setup}.}
\end{enumerate}

\item If you are using existing assets (e.g., code, data, models) or curating/releasing new assets...
\begin{enumerate}
  \item If your work uses existing assets, did you cite the creators?
    \answerYes{Our work uses existing model architectures and ASR methodologies (see Section~\ref{sec:baseline_model}) which we cite.}
  \item Did you mention the license of the assets?
    \answerYes{See Appendix~\ref{app:datasheet}, which mentions that the code and dataset will be released under CC-BY-NC-SA 4.0 license.}
  \item Did you include any new assets either in the supplemental material or as a URL?
    \answerYes{The primary assets are the dataset and the code to reproduce experiments, which can be accessed at \href{https://github.com/facebookresearch/emg2qwerty}{github.com/facebookresearch/emg2qwerty}.}
  \item Did you discuss whether and how consent was obtained from people whose data you're using/curating?
    \answerYes{See Section~\ref{sec:data} which discusses IRB approval for the study and informed consent by the participants.}
  \item Did you discuss whether the data you are using/curating contains personally identifiable information or offensive content?
    \answerYes{We discuss filtering out offensive text from data collection prompts and removing participants' personally identifiable information in Section~\ref{sec:data}.} 
\end{enumerate}

\item If you used crowdsourcing or conducted research with human subjects...
\begin{enumerate}
  \item Did you include the full text of instructions given to participants and screenshots, if applicable?
    \answerNA{No additional instructions were given to the participants beyond what is discussed in Section~\ref{sec:data} and Appendix~\ref{app:datasheet}.}
  \item Did you describe any potential participant risks, with links to Institutional Review Board (IRB) approvals, if applicable?
    \answerYes{The study was approved by an independent external IRB and conducted after consent by the participants as discussed in Section~\ref{sec:data} and Appendix~\ref{app:datasheet}.}
  \item Did you include the estimated hourly wage paid to participants and the total amount spent on participant compensation?
    \answerNA{External participants were compensated pursuant to their agreement with their employer, a third-party vendor company Meta engaged to recruit and hire study participants. See Appendix~\ref{app:datasheet}.}
\end{enumerate}

\end{enumerate}


\newpage

\appendix

\section{Datasheet}
\label{app:datasheet}

\textbf{Motivation:} The motivation for \emgtwoqwerty is to address the lack of wide-spread, sufficiently large, non-invasive surface electromyographic (sEMG) datasets with high-quality ground-truth annotations for a concrete task. sEMG as a technology has the potential to revolutionize how humans interact with devices, and this public dataset is motivated to facilitate progress in this niche domain without needing specialized hardware. This dataset was created by the CTRL-labs group within Reality Labs at Meta.

\textbf{Composition:} The dataset consists of 1,135 HDF5 files, each containing a single session's data. A session here refers to a span of 10 to 45 minutes wherein a participant wearing a sEMG band (see Section~\ref{sec:data}) on each wrist types out prompted text on a keyboard. Each session file includes sEMG data from the left and right bands, the prompted text, the ground-truth key presses recorded by a keylogger, as well as the timestamps for all of these. The sEMG signal is sampled at 2~kHz, and each wristband has 16 electrode channels. The session files include the raw signal after having been high-pass filtered, and after aligning the signals from the left and right wristbands to the nearest timestamps. The 1,135 session files are from a pool of 108 participants involved in the data collection process. Additionally, the dataset includes a metadata file in CSV format to act as an index for the dataset. All metadata have been de-identified to remove any personally identifiable information and does not identify any sub-population. See Section~\ref{sec:data} for additional details on the dataset and Table~\ref{table:dataset_details} for statistics about the dataset such as the number of participants, the total duration, as well as the number of sessions, their duration, and the typing rate broken down along various axes. The recommended data split, which is also what we use in our benchmarks, is to hold out a small number of subjects for personalization, and further split the sessions belonging to each subject for training, validation and testing. This is detailed in Section~\ref{sec:data}. The configuration for the precise data splits used in our experiments is included in our public GitHub repository, together with a script to regenerate them based on a random seed.

\textbf{Collection Process:} The study was approved by an independent Institutional Review Board (IRB), participation was purely on a voluntary basis, participants provided informed consent to include their data, and the data have been de-identified by removing any personally identifiable metadata. External participants were compensated pursuant to their agreement with their employer, a vendor company that Meta engaged to recruit and hire study participants. Participants are further allowed to withdraw from the study during their participation. During a single session of data collection, the participant wears two sEMG wristbands, one on each arm, and types out a series of prompted text on an Apple Magic Keyboard (US English). The text prompts consist of words sampled randomly from a dictionary as well as sentences sampled from English Wikipedia. They were filtered to remove offensive terms, and were processed to only contain lower-case and basic punctuation characters. The sEMG devices are connected via Bluetooth to the laptop that the subject is typing on. A keylogger records the typed characters, together with precise key-down and key-up timestamps. Participants are allowed to freely use the backspace key to correct for typos.

\textbf{Preprocessing/Cleaning/Labeling:} The raw sEMG signals are included in the dataset after high-pass filtering and no further preprocessing was done. Basic quality checks were performed on all the sessions, and those with issues such as missing sEMG due to Bluetooth failures or missing keylogger ground-truth were discarded. Further quality checks were applied to detect artifacts in the signal due to contact with external objects (such as the desk) or unexpected noise, and we include the output of these quality checks per session in the metadata CSV file included in the dataset. No additional labeling was performed beyond the ground-truth provided by the keylogger as part of the collection process.

\textbf{Uses:} The dataset and the associated tooling are meant to be used only to advance sEMG-based research topics of interest within the academic community for purely non-commercial purposes and applications. Our code for baseline models, built on top of frameworks such as PyTorch, PyTorch Lightning and Hydra, is designed such that it can be easily extended to the exploration of different models and novel techniques for this task. The dataset and the associated code are not intended to be used in conjunction with any other data types.

\textbf{Distribution and Maintenance:} The dataset and the code to reproduce the baselines can be accessed at \href{https://github.com/facebookresearch/emg2qwerty}{github.com/facebookresearch/emg2qwerty}. The dataset is hosted on Amazon S3, the code to reproduce the baseline experiments on GitHub, and they are released under CC-BY-NC-SA 4.0 license. We welcome contributions from the research community. Any future update, as well as ongoing maintenance including tracking and resolving issues identified by the broader community, will be performed and distributed through the GitHub repository.

\section{Ethical and Societal Implications}
\label{app:ethics}
Our dataset contains de-identified timeseries of sEMG recorded from consenting participants while touch typing prompted text on a keyboard, together with key-press sequences and timestamps recorded by keylogger to act as ground-truth. The sEMG and key-press data being made available do not provide the ability to identify individuals who participated in the research study.


Given that the amount of data collected per subject is variable among the 108 participants as noted in Table~\ref{table:dataset_details}, a generic model trained from the dataset could end up performing better for a subset of the population. While we release all available data to facilitate research, for experiments that demand keeping the amount of data per subject constant, we remark that n=94 out of the 108 participants in the dataset have 2 hours or more data each.

Beyond this paper, the broader usage of sEMG, and the specific development of sEMG-based textual input models, may pose novel ethical and societal considerations, while also offering numerous societal benefits. sEMG allows one to directly interface a person's neuromotor intent with a computing device. This can be used to create novel device controls for the general population as well as facilitate more inclusive human-computer interactions (HCI) for people who might struggle to use existing interfaces such as those with tremor.

\section{Dataset and Code Access}

The entirety of the \emgtwoqwerty dataset can be downloaded from \url{https://fb-ctrl-oss.s3.amazonaws.com/emg2qwerty/emg2qwerty-data-2021-08.tar.gz}. The dataset consists of 1,136 files in total---1,135 session files spanning 108 users and 346 hours of recording, and one \texttt{metadata.csv} file. Each session file is in a simple HDF5 format and includes the left and right sEMG signal data, prompted text, keylogger ground-truth, and their corresponding timestamps.

The associated code repository to load the dataset and reproduce the experiments in Section~\ref{sec:baselines} can be found at \url{https://github.com/facebookresearch/emg2qwerty}. The \texttt{README.md} file in the GitHub repository contains detailed instructions for installing the package and running experiments with the precise model configuration and hyperparameters needed to reproduce the baseline experimental results in Table~\ref{table:experimental_results}.

Model checkpoint files for the experimental results, as well as the 6-gram character-level language model used, can be found at \url{https://github.com/facebookresearch/emg2qwerty/tree/main/models}. The ``Testing'' section of \texttt{README.md} contains the commands necessary to reproduce the numbers in Table~\ref{table:experimental_results} using these checkpoint files.

The Hydra configuration for the dataset splits can be found under the \texttt{emg2qwerty/config/user} directory in the GitHub repository, and contains the training, validation and test splits for the user generalization and personalization benchmarks discussed in Section~\ref{sec:data}. They can be re-generated by running the Python script \texttt{emg2qwerty/scripts/generate\_splits.py}. Detailed statistics about the \emgtwoqwerty dataset beyond what is reported in Table~\ref{table:dataset_details} can be generated by running the Python script \texttt{emg2qwerty/scripts/print\_dataset\_stats.py}.

The dataset and the code are CC-BY-NC-SA 4.0 licensed, as found in the \texttt{emg2qwerty/LICENSE} file, and will continue to be maintained via GitHub.

\end{document}